%% file: main.tex
\pdfoutput=1
\documentclass[sigconf,screen,screen,authorversion]{acmart}
\usepackage{fancyhdr}
\AtBeginDocument{%
    \addtolength{\footskip}{2.0\baselineskip}%
    \fancyfoot[L]{\textit{\textit{This version is a pre-print. For the definitive Version of Record, please refer to the ACM Digital Library.}}}%
}

\usepackage{algorithmic}
\usepackage{graphicx}
\usepackage{textcomp}
\usepackage{xcolor}
\newcommand*{\vsepfbox}[1]{%
  \begingroup
    \sbox0{\fbox{#1}}%
    \setlength{\fboxrule}{0pt}%
    \mbox{\kern-\fboxsep\fbox{\unhbox0}\kern-\fboxsep}%
  \endgroup
}

\usepackage{hyperref} 
\input{ResearchQuestions}
\usepackage{fontawesome}

\usepackage{framed}

\definecolor{mygray}{gray}{0.9}

\newenvironment{boxC}{
    \MakeFramed{\hsize\linewidth\advance\hsize-\width\FrameRestore}\noindent%
}{%
    \endMakeFramed
}

\usepackage{tabularx}
\usepackage{multirow}

\newcounter{keyTakeAwaysCounter}

\newenvironment{keyTakeAways}[1][Key Take Away]
    {
    \addtocounter{keyTakeAwaysCounter}{1}
        \begin{boxC}
        \faLightbulbO ~ \thekeyTakeAwaysCounter. \textbf{#1}.\\
        }{
        
        \end{boxC}
}

\def\BibTeX{{\rm B\kern-.05em{\sc i\kern-.025em b}\kern-.08em
    T\kern-.1667em\lower.7ex\hbox{E}\kern-.125emX}}
\copyrightyear{2024}
\acmYear{2024}
\setcopyright{rightsretained}
\acmConference[ESEM '24]{Proceedings of the 18th ACM / IEEE International Symposium on Empirical Software Engineering and Measurement}{October 24--25, 2024}{Barcelona, Spain}
\acmBooktitle{Proceedings of the 18th ACM / IEEE International Symposium on Empirical Software Engineering and Measurement (ESEM '24), October 24--25, 2024, Barcelona, Spain}\acmDOI{10.1145/3674805.3695401}
\acmISBN{979-8-4007-1047-6/24/10}

\begin{document}

\title{Beyond Words: On Large Language Models Actionability in Mission-Critical Risk Analysis}


\input{RealAuthors}

\begin{abstract}
\textbf{Context}. Risk analysis assesses potential risks in specific scenarios. Risk analysis principles are context-less; the same methodology can be applied to a risk connected to health and information technology security. Risk analysis requires a vast knowledge of national and international regulations and standards and is time and effort-intensive. A large language model can quickly summarize information in less time than a human and can be fine-tuned to specific tasks.

\textbf{Aim}. Our empirical study aims to investigate the effectiveness of Retrieval-Augmented Generation and fine-tuned LLM in Risk analysis. To our knowledge, no prior study has explored its capabilities in risk analysis. 

\textbf{Method}.  We manually curated \totalscenarios unique scenarios leading to \totalsamples representative samples from over 50 mission-critical analyses archived by the industrial context team in the last five years. We compared the base GPT-3.5 and GPT-4 models versus their Retrieval-Augmented Generation and fine-tuned counterparts. We employ two human experts as competitors of the models and three other three human experts to review the models and the former human expert's analysis. The reviewers analyzed 5,000 scenario analyses. 

\textbf{Results and Conclusions}. HEs demonstrated higher accuracy, but LLMs are quicker and more actionable. Moreover, our findings show that RAG-assisted LLMs have the lowest hallucination rates, effectively uncovering hidden risks and complementing human expertise. Thus, the choice of model depends on specific needs, with FTMs for accuracy, RAG for hidden risks discovery, and base models for comprehensiveness and actionability. Therefore, experts can leverage LLMs for an effective complementing companion in risk analysis within a condensed timeframe. They can also save costs by averting unnecessary expenses associated with implementing unwarranted countermeasures.
\end{abstract}

\maketitle


\section*{Lay Abstract}
\textit{Risk analysis, vital in health and IT security, involves identifying potential risks in various scenarios using universally applicable methods. However, it demands extensive knowledge of regulations and standards, making it time-consuming. Large language models (LLMs) like GPT-3.5 and GPT-4 can quickly summarize information and be tailored for specific tasks, potentially saving time and effort. We curated unique scenarios from over 50 critical analyses conducted in the past five years, comparing the two LLMs. Two human experts performed risk analyses as the model, and three additional experts reviewed the human and LLMs outputs. Results showed human experts were more accurate, but LLMs were faster and provided actionable insights.  LLMs can thus serve as effective tools to complement human expertise, streamlining risk analysis and potentially reducing costs by avoiding unnecessary measures.}

\section{Introduction}
\input{Sections/intro}

\input{Sections/related}
\section{Methodology}\label{sec:design}
\input{Sections/methodology}
\section{Results}\label{sec:results}
\input{Sections/results}

\section{Discussions}\label{sec:discussions}
\input{Sections/discussion}
\section{Industrial Implications}\label{sec:industrial}
\input{Sections/industrial}

\section{Threats to Validity}\label{sec:threats}
\input{Sections/threats}
\section{Ethical Considerations}\label{sec:ethics}

\input{Sections/ethics}
\section{Conclusions}\label{sec:conclusions}
\input{Sections/conclusions}

\begin{acks}
This work has been partially funded by Business Finland project 6GBridge/6GSoft. 
\end{acks}

\bibliography{main}
\balance
\end{document}

%% file: ResearchQuestions.tex
\newcommand{\totalscenarios}{193 }
\newcommand{\trainingscenarios}{135 }
\newcommand{\testingscenarios}{58 }

\newcommand{\totalsamples}{1283 }
\newcommand{\trainingsamples}{926 }
\newcommand{\onlytrainingsamples}{648 }
\newcommand{\validationsamples}{278 }
\newcommand{\testingsamples}{357 }

\newcommand{\rqone}{\texorpdfstring{RQ\textsubscript{1} Do LLMs generate accurate RA?}{RQ1 Do LLMs generate accurate RA?}} 
\newcommand{\rqtwo}{\texorpdfstring{RQ\textsubscript{2} Do LLMs generate actionable and comprehensive RA?}{RQ2 Does LLM generate actionable and comprehensive RA?}}
\newcommand{\rqthree}{\texorpdfstring{RQ\textsubscript{3} Do LLMs outperform human experts in RA?}{RQ3 Do LLMs outperform human experts in RA?}}

\newcommand{\rqoneResult}{\texorpdfstring{Accuracy of LLMs (RQ\textsubscript{1})}{Impact of RAG on LLMs Accuracy (RQ1)}} 
\newcommand{\rqtwoResult}{\texorpdfstring{Actionability of LLMs RA (RQ\textsubscript{2})}{Actionability of LLMs RA (RQ2)}}
\newcommand{\rqthreeResult}{\texorpdfstring{Comparing LLMs with Human Experts (RQ\textsubscript{3})}{Comparing LLMs with Human Experts (RQ3)}}

%% file: RealAuthors.tex





\author{
Matteo Esposito
}
\orcid{0000-0002-8451-3668}
\affiliation{
  \institution{University of Oulu}
  \city{Oulu}
  \country{Finland}
}
\additionalaffiliation{
     \institution{University of Rome Tor Vergata, Rome, Italy} 
     }
\email{matteo.esposito@oulu.fi}

\author{Francesco Palagiano}
\affiliation{
\institution{Multitel}
  \city{Rome}
  \country{Italy}
}
\additionalaffiliation{
     \institution{Multitel di Lerede  Alessandro \& C. s.a.s.} 
     }
    \email{f.palagiano@multitelsrl.it}
\orcid{0009-0006-5613-3445}

\author{Valentina Lenarduzzi}
\affiliation{
  \institution{University of Oulu}
  \city{Oulu}
  \country{Finland}
}
\orcid{0000-0003-0511-5133}
\email{valentina.lenarduzzi@oulu.fi}

\author{Davide Taibi}
\affiliation{
  \institution{University of Oulu}
  \city{Oulu}
  \country{Finland}
}
\orcid{0000-0002-3210-3990}
\email{davide.taibi@oulu.fi}

%% file: Sections/intro.tex
In Information Technology (IT) security, organizations face the challenge of safeguarding digital assets against evolving threats \cite{esposito2023can,cooray2013proactive,esposito2024extensive,sikandar2022context,esposito2023uncovering}. To address the challenges, national governments and international entities, including the ISO, offer guidelines and regulations tailored to quality and risk analysis in mission-critical contexts (MCC), emphasizing the importance of ensuring the success and reliability of operations \cite{cooray2013proactive,esposito2024validate}.

Risk analysis (RA) focuses on grasping the nature of risk and its characteristics, including, where appropriate, the risk severity level. RA involves a deep understanding of uncertainties, risk sources, consequences, likelihood of risk being exploited, events, scenarios, context controls, and the remediations' effectiveness \cite{iso31000}. 






More specifically, RA in software engineering (SE) covers a variety of factors ranging from identifying system vulnerabilities to measuring the impact on stakeholders \cite{charette1989software,verdon2004risk}. It evaluates project scope, technical complexity, resource constraints, and external dependency. This analysis also involves assessing the probability and consequences of identified risks, promoting informed decision-making and effective risk mitigation \cite{kontio2001software,bennett1996risk}. Combining digital and physical elements presents unique challenges, and risk analysis is particularly complex in cyber-physical systems. 

Therefore, we focus beyond the typical software engineering concerns, including physical infrastructure, human factors, and environmental dynamics. To study the risks of cyber-physical systems, we need to assess potential failures through interconnected digital and physical components and consider their cascading effects \cite{esposito2024leveraging}. 

Large Language Models (LLMs) process and produce text resembling human language and are trained on vast amounts of text to perform language tasks like translation, summarisation, question-answering, and text completion \cite{chang2023survey}. Therefore, LLMs are potentially valuable for RA.

We designed our work according to the study performed by Esposito et al. \cite{esposito2024leveraging} that investigated the usage of LLM for preliminary security risk analysis (PSRA). Their findings show that a fine-tuned model (FTM) outperformed the baseline and six human experts regarding proficiency and analysis time in PSRA. PSRA is narrower in scope; on the other hand, our study further explores LLMs' accuracy and actionability in performing a comprehensive RA.

Our study investigated the proficiency, in terms of accuracy and actionability, of LLMs, in identifying vulnerabilities and suggested remediation strategies referring to specific standards and national laws. 

We base our findings on the experience and data acquired during late autumn 2023 and early spring 2024. We leveraged FTM and RAG-assisted models. Notably, HEs demonstrated higher accuracy, while LLMs are quicker and more actionable. Moreover, our findings show that RAG-assisted LLMs have the lowest hallucination rates, effectively uncovering hidden risks and complementing human expertise. Thus, the choice of model depends on specific needs, with FTMs for accuracy and RAG models for comprehensiveness and actionability. Therefore, experts can focus on more comprehensive risk analysis, leveraging LLMs for a practical complementing companion in risk analysis within a condensed timeframe. They can also save costs by averting unnecessary expenses associated with implementing unwarranted countermeasures. Our study's main contributions are as follows:
\begin{itemize}
    \item we perform the first analysis on the impact of RAG and fine-tuning (FT) on LLMs proficiency regarding accuracy and actionability in MCC-RA.
    \item we perform the first comparison between human experts and LLM in MCC-RA.
    \item we provide the first definition of explainability for LLM in MCC-RA.
    \item we define hidden risks in RA and evaluate LLMs and human experts' ability to discover them.
    \item we provide key takeaways for researchers and industrial implications for researchers and practitioners.
    \item we discuss ethical considerations in deploying LLMs in MCC.
\end{itemize}

\textbf{Paper Structure:} Section \ref{sec:background} presents background and definitions for our study whilst Section \ref{sec:related} discusses related works. Section~\ref{sec:design} describes the study design. Section~\ref{sec:results} presents the obtained results, and Section~\ref{sec:discussions} discusses them. Section~\ref{sec:industrial} presents industrial-specific implications while Section~\ref{sec:ethics} discusses ethical implications in deploying LLMs in MCC-RA. Section~\ref{sec:threats} highlights the threats to the validity of our study, and Section~\ref{sec:conclusions} draws the conclusion.

%% file: Sections/related.tex
\section{Background and Definitions}
\label{sec:background}
In this section, we provide essential concepts to our study, encompassing the RAG and fine-tuning procedures and the definitions of vulnerability, threat, scenario, and sample within our specific context.

\subsection{Definitions}
We define ``\textbf{scenario}'' as an excerpt of an RA interview. Likewise, a ``\textbf{sample}'' includes the scenario along with the description of the associated risk, if any. According to \citet{PCM_ANS_TI_002}, we define threat (in Italian ``\textit{minaccia}'') as  ``\textit{the possibility of accidental or deliberate compromise of the security of a computer system or network, resulting in the loss of information confidentiality, data modification/destruction, or interruption or alteration of service behavior}'' \cite{PCM_ANS_TI_002}. 

We define vulnerability (in Italian ``\textit{vulnerabilit\'a}'') as ``\textit{a weakness or lack of adequate controls that may lead to or facilitate the realization of a threat, resulting in compromise, damage, and/or inability to access the relevant information.}'' \cite{PCM_ANS_TI_002}. 


More specifically, the threat, if present, exists regardless of its realization or not. Referring to Table \ref{tab:SCENARIO_AND_VULNERABILITIES}, if we implement a single sign-on authentication system with third-party API interfaces, the \textbf{threat} of \textbf{identity theft} or \textbf{unauthorized access} to the target system is \textbf{always present}, even if such fraudulent access may never occur. The \textbf{vulnerability}, if present, \textbf{facilitates} the occurrence of the threat: in practice, a vulnerability actualizes a threat and makes it more likely. In other words, vulnerabilities essentially increase the probability of a threat occurring, which is why threat and vulnerability are a ``risk'' as a pair and individually, regarding RA.
For example, if the possibility of intentionally manipulating a response to exploit a memory leak, thereby altering or pilfering the authorization token, is not considered during the code's design and development, the likelihood of successfully executing the associated threat of fraudulent access to the system rises \cite{10.1145/2220352.2220353}.

\subsection{Retrieval-Augmented Generation}
\label{subsec:rag}
\begin{figure}
    \centering
    \includegraphics[width=\columnwidth]{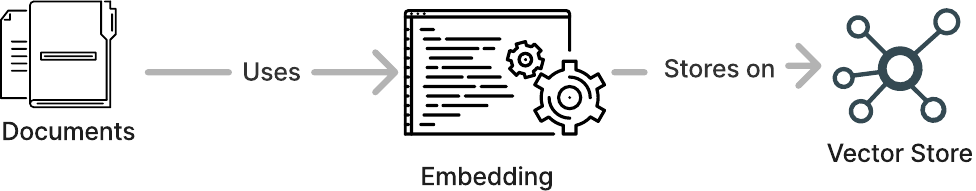}
    \caption{Document Embedding Workflow}
    \label{fig:document_embedding_workflow}
\end{figure}

The knowledge of LLM models is of two types: Parametric and Non-Parametric. It learns parametric knowledge during training, implicitly storing it in the neural network's weights. Conversely, non-parametric knowledge resides in external sources such as a vector database. A vector database stores data as high-dimensional vectors, representing features or attributes mathematically. 
An example of Non-parametric knowledge is the one provided by RAG \cite{lewis2020retrieval} that allows the enhancement of the capabilities of LLMs by using the user question to the model as a query for its vector database and sending back to the model as contextual information, the result of the search and the original user question. The vector database store information as embeddings \cite{lewis2020retrieval}.  Figure   \ref{fig:document_embedding_workflow} presents the embedding workflow. Documents are analyzed through a specialized model (e.g. OpenAi's \textit{text-embedding-3-large} \footnote{\url{https://platform.openai.com/docs/guides/embeddings/embedding-models}}), and the resulting word embedding are stored in the vector store.

\begin{figure*}
    \centering
    \includegraphics[width=2\columnwidth]{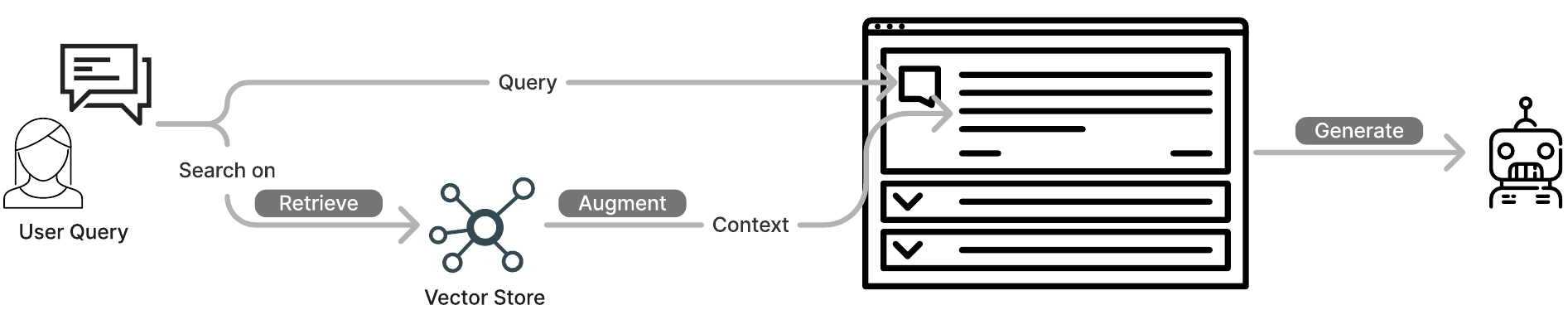}
    \caption{RAG Workflow}
    \label{fig:rag_workflow}
\end{figure*}

Figure \ref{fig:rag_workflow} illustrates the RAG workflow. RAG comprises three phases. \textbf{Retrieve}: The system utilizes the user query to fetch pertinent context from an external knowledge source. To achieve this, the embedding model embeds the user query into the same vector space as the additional context in the vector database. RAG performs a similarity search, yielding the top k closest data objects from the vector database. \textbf{Augment}: The system incorporates the user query and the retrieved additional context into a prompt template. \textbf{Generate}: The retrieval-augmented prompt is sent to the LLM.

\subsection{Model Fine-Tuning}
OpenAI API's documentation states that fine-tuning requires providing conversation samples in a JSONL-formatted document\footnote{\url{https://platform.openai.com/docs/guides/fine-tuning}}. When fine-tuning a model, three roles are available: the \textit{system} role guides the model on how to behave or respond. In our case, we directed the model to reply in a JSON-like output to facilitate automated analysis. We asked the model to identify and link the threats to the national law defining the threats taxonomy. Moreover, the model can ask for more information if the user provided insufficient information for a proper RA. The \textit{user} role simulates the end user asking a question.
Similarly to how we label instances in machine learning, the \textit{assistant} role simulates the model to provide examples of correct answers. We take out 70\% of the \totalsamples, for  FT. Out of  the \trainingsamples samples for FT, we used \onlytrainingsamples samples as a training set and \validationsamples as a validation set. We employed OpenAI's GUI to fine-tune the model.
It is worth noticing that GPT4 fine-tuning is \textbf{not currently generally available}. Therefore, as done by \citet{esposito2024leveraging}, we have used only GPT3.5 as a fine-tuned model.

\subsection{Risk Analysis Process}
\begin{figure*}
    \centering
    \includegraphics[width=\linewidth]{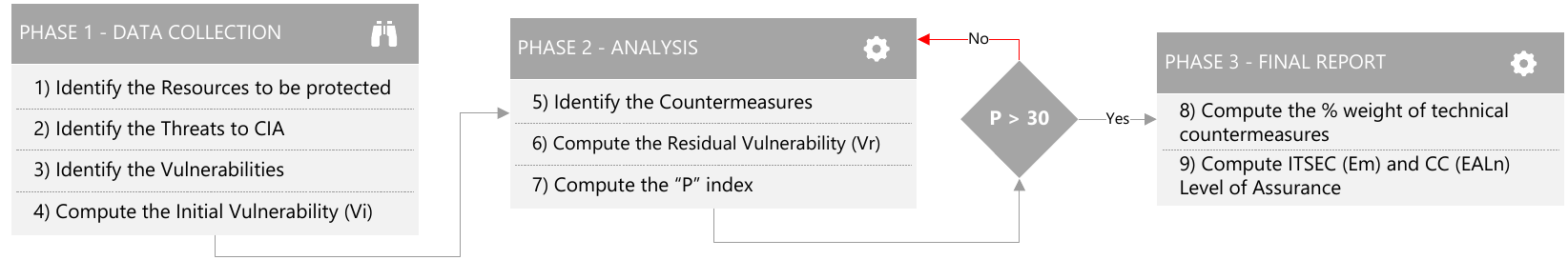}
    \caption{RA Workflow Overview}
    \label{fig:RAOverview}
\end{figure*}

Figure \ref{fig:RAOverview} presents the RA workflow. RA process \cite{PCM_ANS_TI_002} requires acquiring information (Phase 1) regarding the current state of the art about safety and security measures implemented to guard sensitive information, how we select and train the personnel involved, and the procedures adopted in their treatment (Step 1). Such information can be internal documents, best practices, technical blueprints, and interviews.

The personnel must be part of the internal Information Security Department, and a certain number of workers, depending on the dimension of the company or public administration, must be randomly chosen from all the people who treat sensitive information \cite{PCM_ANS_TI_002,iso27001:2022}.

We then assess the data gathered to identify the threats to CIA (Confidentiality, Integrity, Availability)\cite{nist1800-25A} (Step 2) that can be exploited through the vulnerabilities found (Step 3). We assign an initial vulnerability value (Step 4) given the scenario.

In the next step (Phase 2), all the gathered material is analyzed and presented with what the applicable laws, best practices, and technology provide regarding possible countermeasures to the identified threats and vulnerabilities (Step 5).

Following this, we compute the residual vulnerability value, which results after applying all the countermeasures that can be implemented in the system (Step 6).

All of this is to calculate a threshold index (named “P”) that we must at least match to have reasonable confidence that the implemented measures can assure the security of our sensitive information (Step 7).

If the “P” index is equal to or less than 30, we must identify more countermeasures from Step 5. At the same time, if it’s above 30, we can proceed with the last part of the RA process (Phase 3), where we discuss all the findings of the possible countermeasures and compute the percentage weight of technical countermeasures compared to the total number of countermeasures identified (Step 8) (the higher the better \cite{PCM_ANS_TI_002}) to find the equivalent level of information assurance (Step 9), both in regard of ITSEC\cite{PCM_ANS_TI_002,ITSEC1991} and EAL\cite{CC2022} standards.

\section{Related Works}
\label{sec:related}

\citet{hou2023large} conducted a systematic literature review on applying LLMs in SE, aiming to understand their impact and potential limitations. They analyzed 229 research papers from 2017 to 2023. They categorized different LLMs in SE tasks, identifying their features and applications. The authors examined methods for data collection, preprocessing, and application, stressing the importance of curated datasets. Moreover, they investigated strategies for optimizing and evaluating LLM performance in SE. Finally, they identified successful SE tasks where LLMs have been applied. The study discusses current trends, gaps in research, and future study opportunities in LLM for SE.

\citet{sallou2023breaking} address the growing influence of LLMs in SE tasks, highlighting their impact on various aspects such as code completion, test generation, program repair, and code summarization. The paper proposes guidelines for SE researchers and language model providers to address concerns such as closed-source models, data leakage, and reproducibility of findings, with examples of existing good practices and a practical illustration in test case generation within the SE context.



\citet{ni2023chatreport}  developed ChatReport, an LLM-based system, to automate the analysis of corporate sustainability reports. They recognized the challenge of limited resources for human analysis due to the dense and complex nature of these reports, leading to a lack of transparency in sustainability reporting. However, they encountered challenges such as LLM hallucination and the inefficiency of involving domain experts in AI development. 
Our results aligns with previous studies \cite{martino2023knowledge}.

Yuan et al.~\cite{yuan2024rjudge} address the need to benchmark LLMs' behavioral safety in interactive environments. They introduce R-Judge, a benchmark to evaluate LLMs' proficiency in identifying safety risks during agent interactions.  Evaluation of nine LLMs on R-Judge reveals significant room for improvement in risk awareness, with the best-performing model achieving 72.52\% compared to human scores of 89.07\%. Yuan et al.~\cite{yuan2024rjudge} examine the risks of using LLMs as agents in real-world scenarios. In contrast, our work analyzes risks introduced by both environmental factors and human actions. We assess the effectiveness of LLM when it analyzes user-defined scenarios against established laws and standards, and based on this evaluation.

Finally, Esposito et al.~\cite{esposito2024leveraging} investigated the proficiency of LLM in PSRA. Our work deepens Esposito et al.'s~\cite{esposito2024leveraging} investigation, extending the analysis to the more comprehensive RA.Referring to Figure \ref{fig:RAOverview}, the difference between PSRA \cite{esposito2024leveraging} and RA (our work) is that the first one is limited to the first 3 steps, while RA covers the entire process. Therefore, we investigate the comprehensiveness and actionability of the LLM beyond simple accuracy. 

%% file: Sections/methodology.tex
%

This section details the goal and research questions, data collection, and data analysis. Our empirical study follows established guidelines defined by Wohlin et al. \cite{DBLP:books/daglib/0029933}.

The \textit{goal} of our empirical study is to investigate the impact of Retrieve-Augmented Generation (RAG) on the proficiency, in terms of accuracy and actionability, of Large Language Models (LLMs) in the context of Risk Analysis (RA) within mission-critical Information Technology Systems (ITS). Our \textit{perspective} is of both practitioners and researchers seeking to understand whether LLMs are effective tools for quickly performing RA within mission-critical ITS contexts. 

Based on our goal, we defined the following three Research Questions (\textbf{$RQ_s$}). 
\begin{boxC}
\textit{\rqone}  
\end{boxC}

Our ITS context considers a scenario compliant if it does not exhibit threats according to national laws and international standards.
RAG enriches General Purpose LLMs (GPLLMs) output by retrieving authoritative knowledge, such as regulations and ISO standards, and injecting it as context for the user query  \cite{lewis2020retrieval,cai2022recent,mao2020generation} (see Section \ref{subsec:rag}). GPLLM are trained on vast volumes of data and employ billions of parameters to generate original outputs for tasks such as question answering, language translation, and sentence completion. RAG extends LLMs knowledge on specific domains or an organization's internal knowledge base, all without the need for model retraining or fine-tuning \cite{nay2024large,izacard2022atlas}. 

No previous study compared RAG and fine-tuning in reviewing scenarios based on laws and standards in ITS. Our RA samples follow the PCM-ANS TI-002: ``\textit{Security Standard for Military Automatic Data Elaboration (ADE) Systems/Networks}'' by \citet{PCM_ANS_TI_002} which is not freely accessible online or redistributable.
Consequently, it is absent from the model's training data. We are investigating how RAG can improve LLM proficiency by reviewing model outputs against our ground truth. Therefore, we conjectured one \textbf{hypothesis} ($H_1$) as follows:

\begin{itemize} 
    \item $H_{11}$: \textit{There is a significant accuracy difference between LLMs outputs.}
\end{itemize}

\noindent Hence, we defined the \textbf{null hypothesis} ($H_0$) as follows:
\begin{itemize}
    \item $H_{01}$: \textit{There is no significant accuracy difference between LLMs outputs.}
\end{itemize}

We measured the differences in IR accuracy metrics to validate our hypothesis. We look for higher values of Precision and Recall to asses whether a model is more accurate than another. To test the hypothesis, we use the Wilcoxon signed-rank test (WT) as described in Section \ref{subsec:da}.

However, RA is not just about generating reports; it must be actionable. Therefore, we ask:
\begin{boxC}
\textit{\rqtwo}
\end{boxC}
Actionable and comprehensive insights from LLM conducting risk analysis are essential in decision-making processes across various domains. Whether it is assessing financial risks, cybersecurity threats, or environmental hazards, the insights an LLM provides must translate into tangible actions \cite{NEURIPS2023_1190733f}. By transforming complex scenario analysis into clear recommendations, actionable output allows practitioners to implement preventive measures, allocate resources efficiently, and mitigate potential risks proactively. 
Risk analysis remains purely academic without actionable insights, failing to safeguard against emerging threats \cite{aven2015risk}. 

To assess the actionability of the model outputs, we asked three HEs to review the model output. We used Fleis's $\kappa$ to measure the inter-rater agreement (IRA) among the human reviewers (HRs) as described in Section \ref{subsec:da}. 

Based on the reviewer's ratings, we conjectured one \textbf{hypothesis} as follows:

\begin{itemize}
    \item $H_{12}$: \textit{There is a significant difference in the proportion of actionable and comprehensive output between LLMs.}
\end{itemize}

\noindent Hence, we defined the \textbf{null hypothesis} as follows:
\begin{itemize}
    \item $H_{02}$: \textit{There is no significant difference in the proportion of actionable and comprehensive output between the LLMs.}
\end{itemize}

We measured the differences in the proportion of actionable output according to the HRs to validate our hypothesis. More specifically, we define an output as   ``actionable'', if it \textbf{allows a human reader to understand the threat and take action}. Therefore, we reduced the actionability metric to a binary one. We use the WT described in Section \ref{subsec:da} to test the hypothesis.



Finally, LLM has the advantage of analyzing vast amounts of data in less time than humans and storing new information on demand via RAG. Therefore, we ask:
\begin{boxC}
\textit{\rqthree}
\end{boxC}
The level of human expertise and years spent in the field directly impact PSRA quality and proficiency.  We can measure the time required for a new team member to attain the proficiency level of a senior team member in years or even decades. In contrast, LLMs can rapidly assimilate years of training within mere minutes. 
Therefore, it is essential to investigate the performance of the LLM model in terms of accuracy and comprehensiveness with respect to the HEs. Table \ref{tab:experts} provides an overview of the experts' profiles for reviewers and test participants.

Therefore, we conjectured two \textbf{hypotheses}  as follows:
\input{Tables/HUMAN}

\begin{itemize}
    \item $H_{13}$: \textit{LLM models outperforms HE's accuracy.}
    \item $H_{14}$: \textit{The outputs of LLM models are more actionable and comprehensive than a HE's analysis.}
\end{itemize}

\noindent Hence, we defined the \textbf{null hypotheses}  as follows:
\begin{itemize}
    \item $H_{03}$: \textit{There is no significant difference in the accuracy between  LLM models and the HEs.}
    \item $H_{04}$: \textit{There is no significant difference in the proportion of actionable and comprehensive outputs between LLM models and the HEs.}
\end{itemize}

We measure the IR accuracy metrics and the proportion of actionble and comprehensive outputs for HEs and LLM to validate our hypothesis. Therefore, we tested  $H_{03}$ and $H_{04}$ with WT as described in Section \ref{subsec:da}.

\subsection{Study Context}
The context of our case study is an Italian company operating in the civil and military security sector for over 30 years. It is dedicated to researching and developing new technologies for information security and provides products and services aimed at safeguarding data, both at rest and in motion. 

The sample selection derives from previously finalized risk analyses, thus enabling us to obtain the ground truth necessary for fine-tuning and evaluating the model. Table \ref{tab:SCENARIO_AND_VULNERABILITIES} shows a sample from our collection.

\input{Tables/scenarioExample}

We engaged with the company Risk Analysis and Management Team (RAMT) and randomly selected excerpts from the existing finalized documents. To ensure the replicability of our findings, we excluded all sensitive information and samples that could compromise data anonymity. The final dataset comprises \totalsamples samples, encompassing over 50 mission-critical analyses conducted over the last five years. The authors and the RAMT reviewed each sample and agreed unanimously on each classification.

\subsection{Study setup and data collection}
This section delineates our data collection methodology. 
Our study entailed analyzing excerpts from RA interviews. 

The RAMT provided 300 unique scenarios from previously finalized RAs. We excluded 107 scenarios due to sensitive data, representativity concerns, and the need for data anonymity. Regarding representativity, our goal was to fine-tune the model with diverse scenarios, avoiding anchoring towards specific keywords or scenario descriptions. The final dataset comprised \totalsamples samples (\totalscenarios unique scenarios), with \trainingsamples samples (\trainingscenarios unique scenarios) allocated for training and validation and the remaining \testingsamples samples (\testingscenarios unique scenarios) designated for testing. 

We tested two base model, \textit{gpt3.5-turbo-1156} and \textit{gpt4-turbo-preview}, and their RAG-assisted and fine-tuned counterparts. 

We tasked three human experts (HEs) to perform an RA analysis on the same \testingsamples samples we used for the model analysis. Moreover, We tasked three more human experts to review (HRs) the model and the HEs outputs (See subsection \ref{subsec:da} for details). 

\subsection{Data Analysis}
\label{subsec:da}
This section presents our data analysis procedure to address our research questions.

Before explaining how we intend to measure the RQs, we must clarify how the answers from reviewers and models to the different scenarios were validated. Given a specific scenario, HE and models, while analyzing it, can report different identified vulnerabilities or threats that may align or may not align with the ground truth (GT). Nonetheless, GT served as a starting point, but upon considering the answer from HE and models, HR could decide to expand the ground truth on a scenario basis. We decided to allow expanding the GT because, in real-world scenarios, experts can have varying opinions, and within the realm of risk analysis, there is no universal GT. For instance, the ground truth of Multitel is considered ``finalized'' because it was created by expert reviewers and accepted by the client, yet a risk could have been overlooked.

Therefore, we define as \textbf{True Positives} a vulnerability/threat detected that was part of the GT or if the HR decided that the ground truth needed to be extended.

Hence, we define a vulnerability/threat as \textbf{True Negative}, if, and only if, the expert or the model identifies a scenario as not vulnerable or without threats and was indeed not vulnerable or without threats in the GT or the extended GT. 

The HR unanimously agree on the definitive version of the GT before evaluating the answers.

To answer $RQ_1$, we measured the accuracy of the four models on the \testingsamples samples provided.  We adopted standard accuracy metrics in IR as suggested by \cite{esposito2024leveraging,chang2023survey}. We tested $H_{01}$ to assess whether using RAG improves the LLM accuracy and proficiency by comparing their outputs against our GT.
 


To answer $RQ_2$, we measured the actionability and comprehensiveness of the model's RA. Although GT comes from finalized RAs, we considered the possibility for human experts to overlook some details, thus failing to identify specific or more subtle threats. We call this threat ``hidden risks''. Therefore, we also evaluated the ability of models and human experts to extend the ground truth (GT) to measure if the finalized RAs (GT) were spotless or if hidden risks were present. Table \ref{tab:evaluation_criteria} presents the metrics we used to evaluate the scenarios. More specifically, we instructed the HRs to analyze the HE and the model's output according to Table \ref{tab:evaluation_criteria}.

Researchers recently investigated the pressing need for model actionability and explainability \cite{huang2022language,10.1145/3579363} aiming to measure how far we are from responsible usage of AI \cite{arrieta2020explainable}.



\input{Tables/additional_metrics}

To answer  $RQ_2$, we asked three human experts to review (i.e., HRs) the model outputs according to the scenario metrics. We determined the proportion of actionable and comprehensive output and tested $H_{02}$ to asses whether a specific model produces the most actionable or comprehensive output.

Finally, we answer $RQ_3$ asking three HEs to analyze the same \testingsamples sample analyzed by the models. We measured the accuracy of the four HEs and determined the proportion of actionable and comprehensive output from the HE via the previous three reviewers. We compared the human results with the LLM models according to the test performed in $RQ_1$ and $RQ_2$. We tested $H_{03}$ to asses whether the LLMs outperform HEs in terms of accuracy. Similarly, we tested  $H_{04}$ to asses whether the LLM models produce more actionable and comprehensive output than a HE.

\subsubsection{Hypotheses Testing}
We tested $H_{01}$, $H_{02}$, $H_{03}$, and $H_{04}$ via the Wilcoxon signed-rank test \cite{pub.1102728208}, which is a non-parametric statistical test that compares two related samples or paired data. WT uses the absolute difference between the two observations to classify and then compare the sum of the positive and negative differences. The test statistic is the lowest of both. We selected WT  because the accuracy and scenario metrics were not normally distributed; hence, we used it in place of the paired t-test, which assumes a normal data distribution. We set our $\alpha = 0.05$ due to the small size of the dataset. Finally, we do not report the WT p-value table for $H_{01}$ and $H_{02}$ due to space constraints. The whole table is included within the replication package.

\subsubsection{Inter-Rater Agreement}
We analyze HRs agreements, i.e., inter-rater agreements (IRA) via Fleiss's $\kappa$ \cite{fleiss1971measuring} as suggested established empirical standards \cite{ralph2021empirical}. 
When assigning items to different categories, the kappa statistic is commonly used to evaluate the agreement between raters or classifiers. In our context, we used it to measure the level of IRA between the three HRs who reviewed the output of models and  HEs. When comparing the observed agreement level to the expected agreement level, Fleiss's $\kappa$ provides a metric for the reliability and consistency of the categorizations among two or more reviewers. We opted for Fleiss's $\kappa$ over Cohen's $\kappa$ because the latter is restricted to only two reviewers. Table   \ref{tab:kappa-agreement} presents the interpretation Fleiss's $\kappa$ as suggested by \citet{fleiss1971measuring, 10.1093/ptj/85.3.257}.

\input{Tables/kappa}

\subsection{Replicability}
Our replication package includes a Python notebook importable into Google Colab with fine-tuning data, questionnaire answers, model responses and statistical test tables for $H_01$ and $H_02$. The replication package is available on Zeondo \footnote{\url{https://zenodo.org/doi/10.5281/zenodo.10960013}}.

%% file: Tables/HUMAN.tex
\begin{table*}
\small
\centering
\caption{Expert Profiles}
\label{tab:experts}
\begin{tabular}{rrllllr}
\hline
\textbf{Expert} & \textbf{Experience (Years)} & \textbf{Expertise} & \textbf{Common Role} & \textbf{Level} & \textbf{Age Range} & \textbf{Role} \\
\hline
1 & 35 & IT \& Telecom & Security Secretariat & Employee (Manager) & 51-60 & HR \\
2 & 5 & Logistics & Cipher Operator & Employee (Senior) & 31-40 & HE \\
3 & 45 & IT \& Telecom & Director of Security & Executive (Director) & 81-90 & HR \\
4 & 3 & IT \& Telecom & System Administrator & Employee (Junior) & 21-30 & HE \\
5 & 15 & IT \& Telecom & Security Officer & CEO (Owner) & 41-50 & HR \\
\hline
\end{tabular}
\end{table*}

%% file: Tables/scenarioExample.tex
\begin{table}
\small
\centering
\caption{Sample Scenario}
\label{tab:SCENARIO_AND_VULNERABILITIES}

\begin{tabular}{p{4cm}p{4cm}}
\hline \\
\multicolumn{2}{p{4cm}}{\textbf{Scenario}}                                                                                                                                                     \\
\multicolumn{2}{p{8cm}}{\textit{The secure log-in application featuring single sign-on functionality was developed by our in-house team of expert software programmers. However, prior to release, it underwent no testing procedures.}} \\ \\\hline
\\
\multicolumn{2}{p{4cm}}{\textbf{Analysis}}                                                                                                                                                \\
\multicolumn{2}{p{8cm}}{\textit{The described scenario presents at least one security threat.}}                                               \\                                                 \\
The system is vulnerable to asynchronous attacks.

& The system is vulnerable to queuing access.
                                           \\
\textbf{Threat}: M1 - Queuing access & \textbf{Threat}: M4 - Asynchronous attack
                                   \\
\textbf{Vulnerability}: V8 – Inadequate logical access control
 & \textbf{Vulnerability}: V7 – Untested software applications
                         \\
\textbf{Risk Type:} Real                                                   & \textbf{Risk Type:} Real                                                                        \\ \\\hline
\end{tabular}
\end{table}

%% file: Tables/additional_metrics.tex
\begin{table}[htbp]
\small
    \centering
    \caption{Scenario Metrics}
    \begin{tabular}{@{}p{1.7cm}p{6.7cm}@{}}
    \hline
    Metric &  Description \\ \hline
    Comprehensive & Indicates whether the analysis of a scenario has captured the risk and vulnerabilities associated with it. \\
    Actionable &  Indicates whether the analysis related to each scenario are useful or effective in prompting a specific action. \\
    Extended GT &  Indicates whether at least one analysis has extended or enriched the ground truth of a specific scenario. This occurs when a vulnerability not present in the ground truth is identified, and the human reviewer agrees that it was a legitimate threat. \\
    \hline
    \end{tabular}
    \label{tab:evaluation_criteria}
\end{table}

%% file: Tables/kappa.tex
\begin{table}[tb]
\centering
\small
\caption{Interpretation of $\kappa$ values for measuring agreement}
\label{tab:kappa-agreement}
\begin{tabular}{ll}
\hline
Value of $\kappa$ & Interpretation \\
\hline
$\kappa < 0$ & No agreement \\

$0 \leq \kappa < 0.4$ & Poor agreement \\

$0.4 \leq \kappa < 0.6$ & Discrete agreement \\

$0.6 \leq \kappa < 0.8$ & Good agreement \\

$0.8 \leq \kappa < 1$ & Excellent agreement \\
\hline
\end{tabular}
\end{table}

%% file: Sections/results.tex
This Section presents our study results and answers our RQs.

\subsection{Inter-Rater Agreement}
\begin{figure}
    \centering
    \includegraphics[width=\columnwidth]{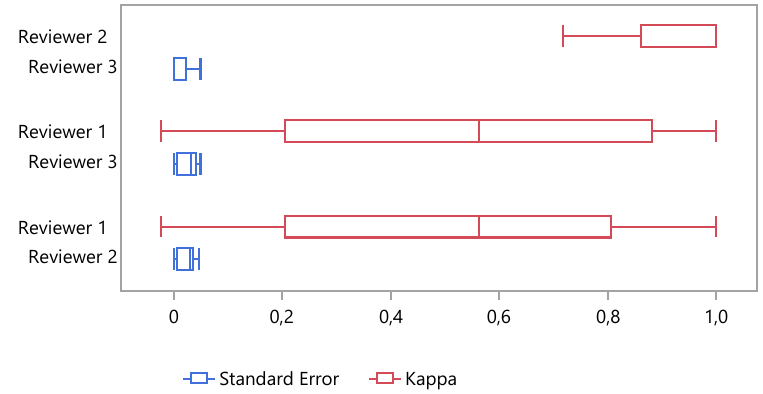}
    \caption{Fleiss's Kappa and Standard Error for IRA}
    \label{fig:IRA}
\end{figure}

Figure \ref{fig:IRA} presents Fleiss's $\kappa$ and the standard error for IRA among the three HRs. According to Figure \ref{fig:IRA} and Table \ref{tab:kappa-agreement} we note that there is a \textbf{strong agreement between HR2 and HR3} while we also note that there is a \textbf{discrete agreement, on average between HR1 and HR3, and HR1 and HR2}.
In all three comparisons, the average standard error is lower than 0,1.
We use IRA to show the differences in HR's reviews. We did not use IRA to reach a consensus as to how RA is conducted in real-world scenarios. RAs are usually reviewed by three or more internal reviewers, and the finalized report is compiled. Our work focused on generating the RA analysis for the reviewers and not for the finalized report, as we envision LLMs as tools aiding HEs and not as their replacements.

\subsection{\rqoneResult}
Table \ref{tab:RQ1} shows the average IR accuracy metrics of the five LLMs. According to Table \ref{tab:RQ1}, models exhibit adequate Accuracy, Precision, and F1 with values between 0.5 and 0.6 for all models except for GPT 3.5 FT. Nonetheless, \textbf{all models show high values for Recall, above 0.8}. 

We used all-pair WT to test $H_{01}$, and we can reject the null hypothesis for all the pairs and metrics. Therefore, we can affirm that \textbf{there is a statistically significant difference in the accuracy of the models}. All the p-values were statistically significant. (See the replication package for WT results.)

Table \ref{tab:comparison_gain} presents the gains comparing each model. According to Table \ref{tab:RQ1}  and Table \ref{tab:comparison_gain} the fine-tuned GPT3.5 model is the most accurate among LLMs in terms of accuracy, F1 Score, and Precision; on the other hand, the RAG-assisted GPT3.5 model exhibit perfect Recall. Considering our findings, we can affirm that \textbf{LLMs can generate accurate RA.} 
\input{Tables/RESULTS/RQ1}
\input{Tables/RESULTS/RQ1-GAIN}


\subsection{\rqtwoResult}
Table \ref{tab:RQ2} shows the percentage of testing scenario analysis made actionable, comprehensive, or their ground truth extended by the models and the percentage of scenario accuracy metrics of the five LLMs. According to Table \ref{tab:RQ2}, models reliably produce actionable and comprehensive scenario analysis. Moreover, we note that \textbf{RAG-based model is more prone to extend the GT}, while base models produce more actionable and comprehensive analysis. We note that FTM did not produce a single actionable or comprehensive scenario analysis or extend the GT. Moreover, we note that each message deemed actionable was also labeled as comprehensive. Therefore we can affirm that \textbf{LLMs can generate actionable and comprehensive RA}.

\input{Tables/RESULTS/RQ2}
We used all-pair WT to test $H_{02}$, and we can not reject the null hypothesis for all the pairs and metrics. Therefore, we must affirm that \textbf{there is not a statistically significant difference in the actionability and comprehensiveness of the models}. (See the replication package for WT results.)

\subsection{\rqthreeResult}
Table \ref{tab:LLM_HEs_acc} shows the average IR accuracy metrics for the five models and the HEs. According to Table \ref{tab:LLM_HEs_acc}, models exhibit similar accuracy as the HEs with slightly lower values for all models except for GPT 3.5 FT. Nonetheless, \textbf{all models show a higher Recall than HEs}. 

Table \ref{tab:H03} shows WT p-values for $H_{03}$. According to Table \ref{tab:H03}, we can reject the null hypothesis for Accuracy, F1 Score, and Precision but not for Recall. Therefore, we can affirm that \textbf{there are significant differences in the accuracy between HEs and LLMs}. More specifically, according to Table \ref{tab:LLM_HEs_acc}  and Table \ref{tab:H03} we note that \textbf{humans are more accurate than LLMs}.

\input{Tables/RESULTS/RQ3.1}
\input{Tables/WT/H03}

Table \ref{tab:LLM_HEs_scen} shows the percentage of testing scenario analysis made comprehensive, actionable or their ground truth extended by the models or the HEs. According to Table \ref{tab:LLM_HEs_scen} we note \textbf{that models produce reliably more comprehensive scenario analysis than HEs}. Moreover, HEs never extended GT. Therefore \textbf{LLMs can discover risks humans couldn't}.

Regarding $H_{04}$, we can reject the null hypothesis for the metrics Extended GT (p-value 0.0115, gain 28\%)  and Actionable (p-value 0.0031, gain 65\%). Therefore, we can affirm that \textbf{the output of the LLMs model is more actionable than HEs analyses}. While we cannot reject it for the Comprehensive metrics (p-value 0.2312, gain 12\%).

\input{Tables/RESULTS/RQ3.2}

Finally, Table \ref{tab:time} shows the time spent on the review or analysis by humans and models. The HRs spent, on average, 38 minutes reviewing a single RA analysis. On the other hand, \textbf{the advantage of LLMs on HEs is evident.} On average, LLMs spent 13.6 minutes to perform RA analysis. On the other hand, the HEs spent approximately 179,5 minutes, i.e., 3 hours, to perform a single RA analysis.

Finally, considering our findings, we can affirm that \textbf{LLMs can outperform humans in actionability, hidden risk discovery, and analysis speed}. Nonetheless, LLMs fall short in accuracy. Hence, future works must focus on improving LLMs' RA accuracy.
\input{Tables/RESULTS/time}

%% file: Tables/RESULTS/RQ1.tex
\begin{table}[]
\centering
\caption{LLM IR Accuracy Metrics}
\label{tab:RQ1}

\begin{tabular}{lllll}
\hline
Model        & Accuracy & F1 Score & Precision & Recall   \\ \hline
GPT3.5       & 0.484375 & 0.611765 & 0.45614   & 0.928571 \\
GPT3.5 + FT  & 0.827848 & 0.89881  & 0.827397  & 0.983713 \\
GPT3.5 + RAG & 0.467949 & 0.637555 & 0.467949  & 1        \\
GPT4         & 0.650794 & 0.685714 & 0.571429  & 0.857143 \\
GPT4 + RAG   & 0.552941 & 0.634615 & 0.485294  & 0.916667 \\ \hline
\end{tabular}
\end{table}

%% file: Tables/RESULTS/RQ1-GAIN.tex
\begin{table}
    \centering
    \caption{Gain for each comparison pair}
\resizebox{\columnwidth}{!}{%
    \begin{tabular}{m{1.6cm}m{1.6cm}llll}
    \hline
    \multicolumn{2}{c}{Comparison}  & Accuracy  & F1 Score  & Precision  & Recall  \\
    \hline
    GPT3.5-FT     & GPT3.5         & +0.343473     & +0.287045     & +0.371257      & +0.055142   \\
    GPT3.5-RAG    & GPT3.5         & -0.016426     & +0.02579      & +0.011809      & +0.071429   \\
    GPT4            & GPT3.5         & +0.166419     & +0.073949     & +0.115289      & -0.071428   \\
    GPT4-RAG      & GPT3.5         & +0.068566     & +0.02285      & +0.029154      & -0.011904   \\
    GPT3.5-FT     & GPT3.5-RAG   & +0.359899     & +0.26121      & +0.359448      & -0.071429   \\
    GPT3.5-FT     & GPT4           & +0.177054     & +0.287045     & +0.371257      & +0.071428   \\
    GPT3.5-FT     & GPT4-RAG     & +0.275907     & +0.264145     & +0.342103      & +0.071428   \\
    GPT3.5-RAG    & GPT4           & +0.182743     & +0.02385      & +0.106719      & +0.142857   \\
    GPT3.5-RAG    & GPT4-RAG     & +0.08479      & -0.00214      & -0.018855      & +0.071429   \\
    GPT4            & GPT4-RAG     & -0.097853     & -0.051099     & -0.086165      & +0.071429   \\
    \hline
    \end{tabular}%
    }
    \label{tab:comparison_gain}
\end{table}

%% file: Tables/RESULTS/RQ2.tex
\begin{table}[]
\centering
\caption{LLM Scenario Metrics}
\label{tab:RQ2}

\begin{tabular}{llll}
\hline
Model        & Actionable & Comprehensive & Extended GT    \\ \hline
GPT3.5       & 0.827586   & 0.827586      & 0.091954     \\
GPT3.5 + RAG & 0.672414   & 0.672414      & 0.666667     \\
GPT3.5 + FT  & 0          & 0             & 0           \\
GPT4         & 0.91954    & 0.91954       & 0.264368    \\
GPT4 + RAG   & 0.850575   & 0.850575      & 0.367816    \\ \hline
\end{tabular}
\end{table}

%% file: Tables/RESULTS/RQ3.1.tex
\begin{table}[]
\centering
\caption{LLM and HEs IR Accuracy Metrics}
\label{tab:LLM_HEs_acc}

\begin{tabular}{lllll}
\hline
          & Accuracy & F1 Score & Precision & Recall   \\ \hline
HUMAN     & 0.856452 & 0.890099 & 0.891884  & 0.891392 \\
LLM       & 0.567584 & 0.648739 & 0.513784  & 0.892857 \\
LLM + FT  & 0.827848 & 0.89881  & 0.827397  & 0.983713 \\
LLM + RAG & 0.510445 & 0.636085 & 0.476621  & 0.958333 \\ \hline
\end{tabular}
\end{table}

%% file: Tables/WT/H03.tex
\begin{table}
\centering
\caption{P-value and gain of the scenario metrics differences, i.e. H$_{03}$ test results.}
\label{tab:H03}
\begin{tabular}{lllll}
\hline
Metric  & Accuracy        & F1 Score        & Precision       & Recall  \\ \hline
p-value & \textit{0,0004} & \textit{0,0047} & \textit{0,0004} & 0,1577  \\ 
gain    & +25.97\%         & +19.64\%         & +33.02\%         & -4.58\% \\\hline
\end{tabular}
\end{table}

%% file: Tables/RESULTS/RQ3.2.tex
\begin{table}[]
\centering
\caption{LLM and HEs Scenario Metrics}
\label{tab:LLM_HEs_scen}

\begin{tabular}{llll}
\hline
          & Actionable & Comprehensive & Extended GT \\ \hline
HUMAN     & 0          & 0.534483      & 0           \\
LLM       & 0.873563   & 0.873563      & 0.178161    \\
LLM + FT  & 0          & 0             & 0           \\
LLM + RAG & 0.761494   & 0.761494      & 0.517241    \\ \hline
\end{tabular}
\end{table}

%% file: Tables/RESULTS/time.tex
\begin{table}[]
\centering
\caption{Time spent on the review or analysis by humans and models}
\label{tab:time}
\resizebox{\columnwidth}{!}{%
\begin{tabular}{lll|lll}
\hline
     & Effort Type & Time (min) &                   & Effort Type & Time (min) \\ \hline
HR 1 & REVIEW      & 50         & GPT3.5 BASE       & ANALYSIS    & 1          \\
HR 2 & REVIEW      & 33         & GPT3.5 RAG        & ANALYSIS    & 15         \\
HR 3 & REVIEW      & 33         & GPT3.5 FT & ANALYSIS    & 11         \\
HE 1  & ANALYSIS    & 211        & GPT4 BASE         & ANALYSIS    & 21         \\
HE 2  & ANALYSIS    & 148        & GPT4 RAG          & ANALYSIS    & 20         \\ \hline
\end{tabular}%
}
\end{table}

%% file: Sections/discussion.tex
This section discusses the implications of the results for researchers and practitioners. We provide take-away messages for each key finding. 

\subsection{Model Accuracy}
 We found that HEs were more accurate than LLMs. Nonetheless, according to Table \ref{tab:time}, LLMs speed was no match for HEs. Moreover, LLMs showed improved Recall scores. In a security context, we tend to favor Recall over Precision as we prefer to avoid false negatives rather than false positives \cite{esposito2024leveraging}

\begin{keyTakeAways}[LLMs as RA copilot]
\textit{LLMs proved to be quicker, more comprehensive, and actionable than humans with a fraction of the time spent. Therefore, considering a trade-off between accuracy and speed, LLMs are a valuable tool in the hands of experts to quickly asses complex scenarios.}
\end{keyTakeAways}

\subsection{Model Explainability}
LLMs are increasingly employed in essential areas like healthcare, finance, and policy. Therefore, we must ensure that domain experts can effectively collaborate with these models \cite{alkhamissi2022review,lakkaraju2022rethinking}. 

We refer to \citet{arrieta2020explainable} definition of explainable AI:
``\textit{Given an audience, an explainable Artificial Intelligence produces details or reasons to make its functioning clear or easy to understand}.''

Explainability serves as a means to connect human decision-makers with machine learning models, helping to bridge the gap between the two \cite{10.1145/3581783.3612817, alkhamissi2022review}.

In our study, we tasked HR to measure the actionability of the LLM's output. According to Table \ref{tab:LLM_HEs_scen}, LLMs provided consistently actionable outputs.  Therefore, according to \citet{arrieta2020explainable}, we define our \textit{audience} as the RA experts, including those who analyze the risk and assess the report. As a \textit{metric} to measure explainability, we use the actionability of the model.

In our context, according to \citet{arrieta2020explainable} definition, an actionable output is self-explanatory regarding the model's inner workings. Specifically, in the RAG-assisted model, the knowledge of acronyms, threats, and vulnerabilities is traceable back to the laws. For instance, the traceability is guaranteed by the \textit{assistant API}\footnote{\url{https://platform.openai.com/docs/assistants/overview}} responds containing annotations that explain what information was taken from which document.
\begin{keyTakeAways}[Actionability as Explainability]
\textit{In a mission-critical risk analysis context, we can use actionability as a proxy metric for measuring model explainability.}
\end{keyTakeAways}

\subsection{LLMs Hallucination}
Although the dataset on which LLMs are trained is vast, it can't be comprehensive in all human knowledge. Hence, LLMs sometimes cheat and makeup facts \cite{ji2023towards,martino2023knowledge,yao2023llm,sevastjanova2022beware}. Our analysis considers hallucination and provides the first measurement of LLM hallucination in RA.

\begin{figure}
    \centering
    \includegraphics[width=\columnwidth]{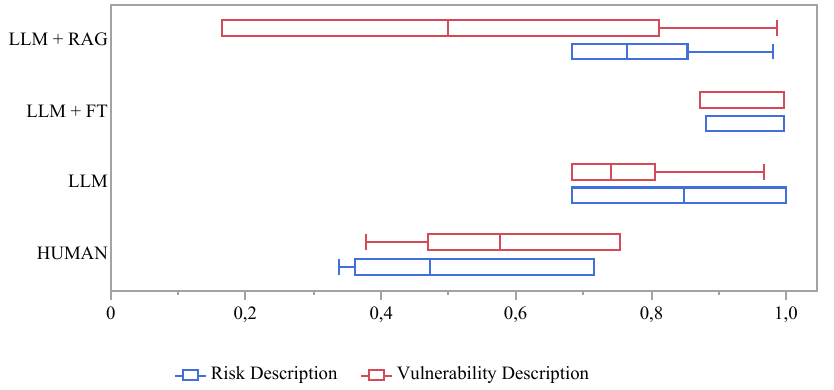}
    \caption{Distribution of the proportion of hallucinated threat and vulnerability description.}
    \label{fig:hallucinations}
\end{figure}

Figure \ref{fig:hallucinations} shows the distribution of the proportion of hallucinated threat and vulnerability description. In our specific context, we define hallucination for a model or a human as an analysis where a specified threat or vulnerability either does not exist in the tables provided by \citet{PCM_ANS_TI_002} or is incorrectly used instead of another. According to Figure \ref{fig:hallucinations}, humans hallucinated, on average, half of the scenarios. It is worth noticing that \textbf{RAG assisted LLMs hallucinate the less}. Our findings align with \citet{martino2023knowledge}. Moreover, RGA-assisted models also challenged HEs in hallucinations on vulnerability description, performing, on average, slightly better than HEs.

\begin{keyTakeAways}[Hallucination and human performance]
\textit{RAG-assisted LLMs reduce hallucinations and can challenge human experts in describing vulnerabilities.}
\end{keyTakeAways}

\subsection{Comparison with Human Experts}
To our knowledge, no one has previously assessed the abilities of LLMs to perform a full scenarios RA.

According to Table \ref{tab:LLM_HEs_scen}, we noted that HEs never extended the GT. Therefore, we can think that specific scenarios hide risks humans cannot grasp. 
\begin{keyTakeAways}[LLMs as a tool for discovering hidden risks]
\textit{In a mission-critical risk analysis context, we should assess each possible risk to the best of our knowledge. Nonetheless, humans can overlook details that LLMs can track down.}
\end{keyTakeAways}
Moreover,  Table \ref{tab:LLM_HEs_scen} also shows that the two models that extended the GT the most were RAG-assisted. Conversely, RAG-assisted GPT4 slightly underperforms in comprehensiveness and actionability.

\begin{keyTakeAways}[To RAG or Fine-Tune, a bards dilemma]
\textit{Choosing the suitable model is essential to ensure hitting the right target. According to our findings, FTM proved to be the most accurate model, while RAG is unmatched in discovering hidden risks and hallucinating the less. In contrast, the base model excels in actionability and comprehensiveness.}
\end{keyTakeAways}

%% file: Sections/industrial.tex
In this section, we discuss industrial-specific implications. The pervasive integration of technology across various domains of human activity, coupled with escalating aggression from attackers driven by fraudulent, demonstrative, or political motives, has resulted in a surge not only in the development of intrusion tools and techniques but also in an exponential rise in the number and diversity of attacks. \cite{clusit2024} The dynamic nature of the scenario clashes with the lengthy execution times of processes like risk analysis, which entail complex steps such as data collection, analysis, evaluation, countermeasure identification, resource allocation, procedural adjustments, and staff training.

Our findings, coupled with \cite{esposito2024leveraging}, highlight LLMs capabilities in at least two of risk analysis tasks:
\begin{enumerate}
    \item the \textbf{preliminary phase} in which the reference context is defined by evaluating which elements of the system constitute a source of risk \cite{esposito2024leveraging};
    \item the \textbf{detailed analysis phase of the sources of risk}, specifying the specific threats and related vulnerabilities that affect the system as done in our work.
\end{enumerate}
Referring to Figure \ref{fig:RAOverview}, aside from the initial step of identifying the resources to protect, our approach, leveraging LLMs, can significantly aid and expedite the subsequent steps of the RA workflow. Applying LLMs to speed up evaluation phases by up to 92\% presents a compelling opportunity. This efficiency improvement, coupled with the model's superior performance achieving a 65\% enhancement in actionability, a 12\% increase in comprehensiveness, and a 28\% extension of the GT compared to HEs, raises two key implications:
\begin{enumerate}
    \item LLMs can\textbf{ support experts in the evaluation of scenarios}, intervening only to amend any macroscopic errors in the model and obtaining, at the same time, an indication of which aspect to intervene in to train it more effectively;
    \item we can \textbf{develop a training process for junior evaluators} incorporating plausible scenarios created by senior experts. This process would assist junior evaluators in comparing their answers with those generated by the model, enabling them to identify and understand their errors. By doing so, they can enhance their skills "in-house" without solely relying on extensive fieldwork, although fieldwork remains valuable and should not be eliminated.
\end{enumerate}
In the future, will be essential to analyze LLMs countermeasures' effectiveness, their applicability concerning the standard and available resources, and how they can be used to support not only decision-makers in the security field but also how to help the human resources office in defining staff training processes.

%% file: Sections/threats.tex
In this section, we discuss the threats to the validity of our case study. We categorized the threats in Construct,  Internal, External, and Conclusion validity following established guidelines \cite{DBLP:books/daglib/0029933}.

\textbf{Construct Validity} concerns how our measurements reflect what we claim to measure \cite{DBLP:books/daglib/0029933}. Our design choices, including our measurement process and data filtering, may impact our results. To address this threat, we based our choice on past studies and well-established guidelines in designing our methodology \cite{DBLP:journals/ese/RunesonH09,Basili1994}. 

\textbf{Internal Validity} is the extent to which an experimental design accurately identifies a cause-and-effect relationship between variables \cite{DBLP:books/daglib/0029933}. Our study relies on \totalsamples samples, which can potentially be biased from the sample selection and the MCC. We addressed this issue by sampling over 50 mission-critical analyses conducted by the industrial context team on different fields, from national security to health and education.

\textbf{External Validity} concerns how the research elements (subjects, artifacts) represent actual elements \cite{DBLP:books/daglib/0029933}. Our case study focused on an Italian company operating in the civil and military security field. The use of the Italian language and this company's specific characteristics may limit the findings' generalizability to other organizations or fields. We addressed this concern similar to the internal validity threats by sampling from over 50 mission-critical risk analyses across various fields. Moreover, we chose a GPLLM with no specific Italian language advantages or restrictions. Given the inherent language capabilities of the model and the context-less nature of RA,  having the samples in Italian should not pose any generalizability issues \cite{ogueji2021small,choudhury2021linguistically}.

\textbf{Conclusion Validity} focuses on how we draw conclusions based on the design of the case study, methodology, and observed results \cite{DBLP:books/daglib/0029933}. Our conclusions rely on the specific accuracy metrics chosen, and there may be other aspects or dimensions of performance that we did not consider. To address this potential limitation, we selected metrics from recent related studies that have faced the challenge of validating LLM proficiencies in specific tasks \cite{esposito2024leveraging,chang2023survey}. Moreover, statistical tests threaten the conclusion's validity regarding the appropriateness of statistical tests and procedures, such as assumption violation, multiple comparisons, and Type I or Type II errors. We address this issue using WT instead of the t-test due to the rejection of the normal data distribution hypothesis. 

%% file: Sections/ethics.tex
In this Section, we mention specific ethical issues that can arise from adopting LLM in RA.  Our world is fast-changing. Until recently, AI used to categorize things, but now it is \textbf{serving humans imagination}.
 Therefore, according to \citet{10372461}, SE researchers and practitioners should not ``look away''. 


\textbf{LLM in Risk Analysis.} The deployment of LLMs in security risk analysis raises ethical concerns, particularly regarding false positives and false negatives. False positives may lead to unwarranted suspicion or accusations, potentially infringing on individuals' rights and privacy. Conversely, false negatives could result in overlooked threats, endangering security. 
To deploy an ethical-aware strategy, we must lay down practices for transparent validation processes, bias mitigation, and accountability mechanisms to minimize these risks and ensure fair and accurate outcomes.

\textbf{LLM as a human assistant.} Our study recognize LLMs as \textbf{assistants} rather than \textbf{human replacements} in security risk analysis. While these models offer valuable insights and efficiency, human oversight remains essential for contextual understanding, critical judgment, and ethical decision-making. LLMs can augment human capabilities, streamline processes, and provide data-driven recommendations. Still, their integration should prioritize collaboration with human experts to maintain accountability, mitigate biases, and uphold ethical standards in security analysis.

%% file: Sections/conclusions.tex
In conclusion, we conducted the first preliminary investigation on LLMs' proficiency in terms of accuracy and actionability on RA. Albeit HEs demonstrated higher accuracy, LLMs proved to be quicker and more actionable, making them valuable for time-sensitive situations. Moreover, our findings show that RAG-assisted LLMs have the lowest hallucination rates. LLMs, especially RAG models, excel in uncovering hidden risks, complementing human expertise. Thus, the choice of model depends on specific needs, with FTMs for accuracy, RAG for hidden risks discovery and fewer hallucinations, and base models for comprehensiveness and actionability. In future works, we will conduct a more extensive large-scale analysis to counter biases of small sample size and deepen the findings by tackling lightweight language models and language-specific models.